\documentclass[authoryear,11pt]{elsarticle}

\usepackage{amssymb}
\usepackage{amsthm, amsfonts}
\usepackage{mathtools}
\usepackage{xcolor}
\usepackage{multicol}
\usepackage{float}
\usepackage{newpxtext,newpxmath}
\usepackage{pgfplots}
\usepackage{booktabs}
\usepackage{multirow}
\usepackage{longtable}
\usepackage{graphicx} 
\usepackage{adjustbox}
\usepackage[a4paper, right=22mm, left=22mm]{geometry}
\usepackage{caption,subcaption}
\usetikzlibrary{intersections,backgrounds,external}
\usepgfplotslibrary{fillbetween}
\pgfplotsset{compat=1.18}

\definecolor{prussianblue}{rgb}{0.0, 0.19, 0.33}
\definecolor{internationalorange}{rgb}{1.0, 0.31, 0.0}
\definecolor{awesome}{rgb}{1.0, 0.13, 0.32}
\definecolor{cadmiumorange}{rgb}{0.93, 0.53, 0.18}
\definecolor{brick}{HTML}{B6321C}
\definecolor{chromeyellow}{rgb}{1.0, 0.65, 0.0}
\definecolor{nblue}{HTML}{006EB8}
\definecolor{amber}{rgb}{1.0, 0.49, 0.0}

\usepackage[
colorlinks=true,        
allcolors = prussianblue,  
urlcolor = internationalorange,
citecolor=awesome
]{hyperref}

\usepackage{lineno}
\usepackage[ruled,vlined, linesnumbered]{algorithm2e}
\usepackage{algpseudocode}
\usepackage{subcaption}

\journal{European Journal of Operational Research}
\usepackage{setspace}
\newcommand{\stdf}[1]{\ \mbox{\scriptsize $\pm$ #1}}

\begin{document}

\begin{frontmatter}



\title{Supplementary Materials to\\Graph Convolutional Branch and Bound}

\author[a,b]{Lorenzo Sciandra \corref{cor1}}
\ead{lorenzo.sciandra@unito.it}
\cortext[cor1]{Corresponding author}
\author[b]{Roberto Esposito}
\author[a]{Andrea Grosso}
\author[a]{Laura Sacerdote}
\author[a]{Cristina Zucca}

\affiliation[a]{organization={Department of Mathematics, University of Turin},
             addressline={Via Carlo Alberto, 10},
             city={Turin},
             postcode={10123},
             state={Italy},
             country={Italy}}
\affiliation[b]{organization={Department of Computer Science, University of Turin},
            addressline={Via Pessinetto, 12},
            city={Turin},
            postcode={10149},
            state={Italy},
            country={Italy}}



\begin{abstract}
This article explores the integration of deep learning models into combinatorial optimization pipelines, specifically targeting NP-hard problems. Traditional exact algorithms for such problems often rely on heuristic criteria to guide the exploration of feasible solutions. In this work, we propose using neural networks to learn informative heuristics—most notably, an optimality score that estimates a solution's proximity to the optimum. This score is used to evaluate nodes within a branch-and-bound framework, enabling a more efficient traversal of the solution space. Focusing on the Traveling Salesman Problem, we introduce Concorde, a state-of-the-art solver, and present a hybrid approach called Graph Convolutional Branch and Bound, which augments it with a graph convolutional neural network trained with a novel unsupervised training strategy that facilitates generalization to graphs of varying sizes without requiring labeled data. Empirical results demonstrate the effectiveness of the proposed method, showing a significant reduction in the number of explored branch-and-bound nodes and overall computational time. Some of the results concerning the use of the 1-tree relaxation are in the supplementary materials.
\end{abstract}




\end{frontmatter}



\section{Introduction}

 In these supplementary materials to the paper~\cite{sciandraGCBB_2026}, named Graph Convolutional Branch and Bound (GCBB), we analyze a classical exact solver for the Traveling Salesman Problem (TSP) and its enhanced neural version. The choice of this solver is related to its simplicity, which allows us to control each step and to recognize the effect of our hybrid version on it. As discussed in the main paper, this preliminary study aimed to explore how a Graph Neural Network (GNN) can be employed to improve an exact TSP solver. Implementing the code\footnote{Code, data, and results are available at the following GitHub repository: \url{https://github.com/LorenzoSciandra/GraphConvolutionalBranchandBound}.} from scratch provided full control over the branch-and-bound process and allowed us to identify potential areas for improving heuristic choices. Several ideas developed for the simple exact TSP solver presented here are then applied in the main paper to extend Concorde. These supplementary materials include only the essential details required to understand the experiments presented here, while all other aspects are discussed in the main paper. To help the reader, we repeat here formulas and some concepts introduced in the main paper if useful.


\section{Traveling Salesman Problem and 1-tree branch and bound}

Given an undirected graph $G(V,E)$ where each vertex $i\in V = \{1,2,\dots,n\}$ represents a city and each edge $e={i,j}\in E$ has an associated travel cost $c_e$ the TSP model writes out as:

\begin{align}
    \min \:f(\boldsymbol x) = \min & \sum_{e\in E} c_ex_e\label{tsp:obj} \\
    \text{subject to}\ & \sum_{e\in\delta(i)} x_e=2 && i\in V\setminus\{n\}\label{tsp:degree}\\
                       & \sum_{e\in E}x_e=n \label{tsp:num-edges}\\
                       & \sum_{e\in E(S)} x_e \leq |S| - 1  && \forall\: S \subset V \setminus\{1\}\label{tsp:subtour} \\
                       &x_e \in \{0,1\} &&\forall e\in E\label{tsp:vars}
\end{align}

A classical exact solver for the TSP is a 1-tree branch and bound. Precisely, following~\cite{HK1} and~\cite{HK2}, a reasonable lower bound for the TSP is obtained on a Lagrangian relaxation on the degree constraints~\eqref{tsp:degree} that ensure that each vertex has exactly two incident edges except for the starting one. The solution to the relaxed problem is known as a 1-tree, and it can be efficiently computed in polynomial time using either Kruskal's or Prim's algorithm. Specifically, it involves a minimum spanning tree over all vertices except the starting one, and then adding the two edges incident to it with the lowest weights. Since in a cycle, each vertex can be the starting vertex, the one that defines the maximum lower bound and so minimizes the initial gap with the optimum is usually chosen. For this work, we implemented a branch and bound algorithm following what is proposed in~\cite{Volgenant} and ~\cite{Valenzuela}: at each node of the branch and bound tree, a fixed number of sub-gradient improvements are done to increase the value of the corresponding lower bound found with the 1-tree relaxation. If the resulting node is not closed, some edge-fixing techniques from~\cite{Fixing} are applied to further restrict, whenever possible, the number of edges that are candidates for the branching step. Only after this step, if the subproblem remains open, branching is performed using the binary branching scheme proposed in~\cite{BranchImp}, which integrates an edge-scoring mechanism with the subgradient optimization procedure. Specifically, when performing the subgradient optimization steps, the optimization process eventually converges to the value of the linear relaxation; in the final iterations, the relaxed solution typically oscillates among a few 1-trees. A fractional score, representing its normalized frequency of appearance in this phase, is assigned to each edge. The edge with the score closest to $0.5$ is then selected for branching. A multi-start nearest-neighbor heuristic~\citep{NearestNeig} is applied at the beginning to generate the initial feasible solution and upper bound.


\section{Graph Neural Networks for the TSP}
\label{sec:GNN}
A probability matrix $\mathbf{P}$ that contains, like the one presented in the main paper, probabilities for each edge to belong to an optimal tour is here obtained through a supervisely pre-trained GNN.

More precisely, the pre-trained GNNs proposed in~\cite{GCNTSP} are defined for graphs of fixed dimension $n=\{20,50,100\}$, and the architecture is implemented as follows.

At first, the input features of vertices and edges are mapped into an embedding space with dimension $h=300$ through learnable linear transformations. Then, $L=30$ layers of message passing with residual connections are applied, where a gating mechanism is used to filter the relevant information in the neighboring vertices through a dense attention map. In the post-processing stage, the edge embeddings $\mathbf{e}_{ij}^L$ of the last layer are fed into a Multi-Layer Perceptron to compute the probability 
of the corresponding edge being in an optimal tour with a sigmoid activation.

For each graph size, the training dataset consists of one million graph instances, where vertex positions are sampled uniformly 
from the $[0,1]\times[0,1]$ square, and distances are computed using the $L_2$ norm. During each training epoch, a subset of $10,000$ 
instances are sampled from the training set, and the model is trained to output the symmetric probability matrix $\mathbf{P}$. 
So, the models were trained end-to-end in a supervised manner by minimizing the cross-entropy loss via gradient descent over these probabilities.
The pre-trained models have been made publicly available by the authors of~\cite{GCNTSP} and are used for this first set of experiments.%
\footnote{Downloadable from the paper 
repository \url{https://github.com/chaitjo/graph-convnet-tsp}}

A remarkable limit of this approach lies in the reliance on supervised training, which requires an extensive dataset of optimally solved 
instances. This actually prevents the effective handling of large TSP instances.


\section{Graph Convolutional Branch and Bound}
The goal of this section is to explain how the 1-tree optimality scores $O_T$, formally defined in the main paper, are integrated into the 1-tree-based branch and bound.
The integration takes place in different parts of the branch and bound 
procedure, defining the Graph Convolutional Branch and Bound.

\subsection{Relaxation at the root node}
The lower bound at the root node of the branching tree is computed 
by generating a 1-tree rooted at a suitable vertex.
Among all the possible vertices, it is usually chosen the one that allows the generation of the 1-tree with the highest possible lower bound.
When different 1-trees provide the same lower bound, we break such ties by choosing the one with the highest expected value $\mathbb{E}[O_T]$.

\subsection{Probabilistic Nearest Neighbor}
Leveraging the probabilistic information of the edges now available, we advocate utilizing an algorithm we term the Probabilistic Nearest Neighbor (PNN) for determining the initial feasible solution. The PNN starts from a specific vertex, and, at each step, chooses the edge with the maximum probability that leads to an unvisited vertex. This process is iteratively applied with all vertices as starting points. The minimum tour among all iterations is then compared with the one obtained through the classical multi-start nearest neighbor. Ultimately, only the best tour reached is regarded as the initial feasible solution.

\subsection{Branching variable selection}
The branching operation is performed by selecting or excluding a certain edge from the solution. We adopt the scoring scheme proposed in~\cite{BranchImp} for selecting the branching edge; additionally, in this selection, we break ties by selecting the edge $(i,j)$ with the highest $p_{ij}$ among the edges with maximum Shutler's score.
The rationale behind this is that if the GCN assigns a high probability to an edge because it deems it likely to be part of an optimal tour, then, among peers with the same Shutler's score, it is sensible to prioritize exploration of this edge first.

\subsection{Branching node selection}
In the branch and bound, when a subproblem with a value lower than the current upper bound generates its children through the branching step, these children are added to a list of open decision nodes. In the classic version of the algorithm, these nodes are ordered based on increasing values of the lower bounds given by the associated relaxations (1-trees), adhering to the standard best-first rule.
In our version of the algorithm, when the difference between the lower bounds of two nodes is below a designated threshold, the node exhibiting a superior expected value of the optimality score $O_T$ is chosen.


\section{Results}
\label{ssec:exact-tsp-results}

The graph instances used for testing our Graph Convolutional Branch and Bound are randomly generated by uniformly sampling their coordinates $(v_{i1}, v_{i2})$ in the $[0,1]\times[0,1]$ square, and with Euclidean distances as defined in the TSPLIB collection.
We chose to sample in this subspace because every graph can be rescaled to fit within the unit square without altering the set of optimal tours. 
Moreover, given that we are tackling an NP-hard problem, it is reasonable to establish a time limit, beyond which the computation is halted. When this happens, the problem instance is deemed unsolved, and its computation metrics are disregarded. In our scenario, a time limit of 10 minutes is set, serving as the exit condition. These experiments were conducted on a server that features an Intel Xeon Cascade Lake processor with 4 cores and two NVIDIA Tesla T4 GPUs, each with 15 GB of memory.

\begin{figure*}[!htbp]
\centering
{
\begin{subfigure}[t]{0.42\textwidth}
    \centering
        \begin{tikzpicture}
        \begin{axis}[
            width=0.95\textwidth,
            xlabel={Time (s)},
            ylabel={Fraction of instances solved},
            grid=both,
            major grid style = {dotted},
            legend style={at={(0.2,0.3)},anchor=west},
            font=\footnotesize
        ]
        \addplot[no markers, color=nblue, draw opacity=0.7, line width=1.2pt] table [x=Time, y=Hybrid,col sep=comma]{20_cumulative_profile.csv};

        \addplot[no markers, color=amber, draw opacity=0.7,line width=1.2pt] table [x=Time, y=Classic,col sep=comma]{20_cumulative_profile.csv};

        \legend{GCBB, BB};

        \end{axis}
    \end{tikzpicture}
    
\end{subfigure}
~ 
\begin{subfigure}[t]{0.42\textwidth}
    \begin{tikzpicture}
        \begin{axis}[
            width=0.95\textwidth,
            xlabel={Time (s)},
            ylabel={Fraction w/ better solution},
            grid=both,
            major grid style = {dotted},
            legend style={at={(0.1,0.3)},anchor=west},
            font=\footnotesize
        ]
        \addplot[no markers, color=nblue, draw opacity=0.7, line width=1.2pt] table [x=Time, y=Hybrid,col sep=comma]{20_performance_profile.csv};

        \addplot[no markers, color=amber, draw opacity=0.7, line width=1.2pt] table [x=Time, y=Classic,col sep=comma]{20_performance_profile.csv};

        \end{axis}
    \end{tikzpicture}
\end{subfigure}
\subcaption{\label{fig:20-profiles} 20 vertices.
}
}
\medskip
{
\begin{subfigure}[t]{0.42\textwidth}
    \centering
        \begin{tikzpicture}
        \begin{axis}[
            width=0.95\textwidth,
            xlabel={Time (s)},
            ylabel={Fraction of instances solved},
            grid=both,
            major grid style = {dotted},
            legend pos=south east,
            font=\footnotesize
        ]
        \addplot[no markers, color=nblue, draw opacity=0.7, line width=1.2pt] table [x=Time, y=Hybrid,col sep=comma]{50_cumulative_profile.csv};

        \addplot[no markers, color=amber, draw opacity=0.7, line width=1.2pt] table [x=Time, y=Classic,col sep=comma]{50_cumulative_profile.csv};

        \end{axis}
    \end{tikzpicture}
    
\end{subfigure}
~ 
\begin{subfigure}[t]{0.42\textwidth}
    \begin{tikzpicture}
        \begin{axis}[
            width=0.95\textwidth,
            xlabel={Time (s)},
            ylabel={Fraction w/ better solution},
            grid=both,
            major grid style = {dotted},
            font=\footnotesize
        ]
        \addplot[no markers, color=nblue,draw opacity=0.7, line width=1.2pt] table [x=Time, y=Hybrid,col sep=comma]{50_performance_profile.csv};

        \addplot[no markers, color=amber,draw opacity=0.7, line width=1.2pt] table [x=Time, y=Classic,col sep=comma]{50_performance_profile.csv};

        \end{axis}
    \end{tikzpicture}
\end{subfigure}
\subcaption{\label{fig:50-profiles} 50 vertices.}
}
{
\begin{subfigure}[t]{0.42\textwidth}
    \centering
        \begin{tikzpicture}
        \begin{axis}[
            width=0.95\textwidth,
            xlabel={Time (s)},
            ylabel={Fraction of instances solved},
            grid=both,
            major grid style = {dotted},
            legend pos=south east,
            font=\footnotesize
        ]
        \addplot[no markers, color=nblue,draw opacity=0.7, line width=1.2pt] table [x=Time, y=Hybrid,col sep=comma]{100_cumulative_profile.csv};

        \addplot[no markers, color=amber,draw opacity=0.7, line width=1.2pt] table [x=Time, y=Classic,col sep=comma]{100_cumulative_profile.csv};

        \end{axis}
    \end{tikzpicture}
    
\end{subfigure}
~ 
\begin{subfigure}[t]{0.42\textwidth}
    \begin{tikzpicture}
        \begin{axis}[
            width=0.95\textwidth,
            xlabel={Time (s)},
            ylabel={Fraction w/ better solution},
            grid=both,
            major grid style = {dotted},
            font=\footnotesize
        ]
        \addplot[no markers, color=nblue,draw opacity=0.7, line width=1.2pt] table [x=Time, y=Hybrid,col sep=comma]{100_performance_profile.csv};

        \addplot[no markers, color=amber,draw opacity=0.7, line width=1.2pt] table [x=Time, y=Classic,col sep=comma]{100_performance_profile.csv};

        \end{axis}
    \end{tikzpicture}
\end{subfigure}
\subcaption{\label{fig:100-profiles} 100 vertices. }
}

\caption{\label{fig:perf_prof_exact} Performances achieved by the {\color{nblue}GCBB} and  {\color{amber}BB} solver grouped in two distinct performance profiles for each graph size where an exact model exists. }

\end{figure*}

\subsection{Comparison metrics}

As previously said, in this first part of our work, we avoided the use of predefined libraries for standard solvers because we want to monitor various metrics about the properties of the GCBB that otherwise would be hidden by the software. In addition to conventional metrics like average solution time or average time to achieve optimality, we consider a few metrics based on the number of created nodes in the branch and bound tree, which is a measure that has the advantage of being independent of the execution environment. The combinatorial explosion of these nodes is indeed at the core of the algorithm's high computational complexity. Specifically, we report:

\begin{itemize}

\item \texttt{Total Time}: the cumulative time required for the entire procedure, encompassing instance file parsing, algorithm execution, result output, and neural network inference, if used;
\item \texttt{BB Time}: the computation time dedicated exclusively to the branch-and-bound phase of the algorithm;
\item \texttt{Time to Best}: the elapsed time for the branch-and-bound algorithm to first discover the solution that is ultimately confirmed as optimal;
\item \texttt{BB Tree Depth}: depth of the generated tree;
\item \texttt{Depth of the optimum}: depth at which the optimum is found;
\item \texttt{Generated BB nodes}: total number of nodes required for completing the search for the optimum;
\item \texttt{Explored BB nodes}: cardinality of the subset of Generated BB nodes necessitating a branching step;
\item \texttt{BB nodes before optimum}: number of nodes generated before reaching the optimum.

\item \texttt{Optimality score} $\:\frac{\mathbb{E}[O_T]}{n}$: measures the goodness of the chosen criterion and so of the overall procedure. Indeed, we expect this value to be close to $1$ for all the 1-trees that represent the optimum of the analyzed problems. See the main paper for a formal definition.
\end{itemize}

\subsection{1-tree results}
We derive the value of the mentioned metrics by averaging them on the solutions obtained on $1000$ instances for all graph sizes. They are presented in \autoref{tab:exact-models}, where optimal values are highlighted in bold. If the disparities in the means are statistically significant according to a coupled Wilcoxon signed-rank test at the 5\% significance level, the values are also underlined.

\begin{table}[t]
    \centering
    \renewcommand{\arraystretch}{1.5}
    \caption{
    The first three rows present the percentages of the total instances solved and how the optimum is found. In the following rows, we report the mean values and their confidence intervals for all the chosen metrics. All these values are calculated when an exact GCN exists and refer to the same set of instances solved by both solvers within 10 minutes. A dash (–) is used to indicate that the corresponding metric is not defined for a given solver.
    }
    \label{tab:exact-models}
    \resizebox{\textwidth}{!}{%
        {\normalsize
        \begin{tabular}{lr@{}l r@{}lr@{}l r@{}l r@{}l r@{}l}
            \toprule
            & \multicolumn{4}{c}{$n=20$} & \multicolumn{4}{c}{$n=50$} & \multicolumn{4}{c}{$n=100$}\\
            & \multicolumn{2}{c}{BB} & \multicolumn{2}{c}{GCBB} & \multicolumn{2}{c}{BB} & \multicolumn{2}{c}{GCBB} & \multicolumn{2}{c}{BB} & \multicolumn{2}{c}{GCBB} \\
            \midrule
            Instances solved & 100.0\% && 100.0\% && 100.0\% && 100.0\% && 95.6\% && 95.6\% &\\
            $\qquad$ with NN & 8.9\% && 1.3\% && 0.0\% && 0.0\% && 0.0\% && 0.0\% &\\
            $\qquad$ with PNN & - && 86.0\% && -&& 79.5\% &&-&& 48.12\% &\\
            Total time (s) & \underline{\textbf{0.01}} & \stdf{0.0} & 1.55 & \stdf{0.0} & \underline{\textbf{1.65}} & \stdf{0.85} & 3.25 & \stdf{0.77} & 51.69 & \stdf{5.91} & \textbf{45.73} & \stdf{5.4}\\
            BB time (s) & 0.0 & \stdf{0.0} &  0.0 & \stdf{0.0} & 1.62 & \stdf{0.85} & \underline{\textbf{1.5}} & \stdf{0.77} & 51.46 & \stdf{5.9} & \underline{\textbf{43.25}} & \stdf{5.39}\\
            Time to Best (s) & 0.0 &\stdf{0.0} & 0.0 & \stdf{0.0} & 0.75 & \stdf{0.36} & \underline{\textbf{0.51}} & \stdf{0.24} & 33.60 & \stdf{4.09} & \underline{\textbf{19.1}} & \stdf{3.21}\\
            BB tree depth  & \underline{\textbf{2.41}} &  \stdf{0.16} &  2.79 &  \stdf{0.19} & 9.16 & \stdf{0.39} &  \underline{\textbf{8.29}} &  \stdf{0.37} & 16.41 &  \stdf{0.4} &  \underline{\textbf{15.54}} & \stdf{0.44}\\
            Depth of the opt. & 1.61 & \stdf{0.08} & \underline{\textbf{0.36}} & \stdf{0.07} & 5.83 & \stdf{0.21} & \underline{\textbf{1.65}} & \stdf{0.22} & 11.03 & \stdf{0.26} & \underline{\textbf{6.44}} & \stdf{0.43}\\
            Generated BB nodes  & \underline{\textbf{7.18}} &  \stdf{1.18} & 7.89 & \stdf{1.21} &  874.98 & \stdf{416.66} & \underline{\textbf{805.14}} & \stdf{389.44} &  4439.60 & \stdf{500.27} &  \underline{\textbf{3590.77}} & \stdf{443.48}\\
            Explored BB nodes & \underline{\textbf{6.34}} & \stdf{1.05} & 7.46 & \stdf{1.02} & 778.19 & \stdf{398.87} & \underline{\textbf{753.88}} & \stdf{376.0} & 3308.89 & \stdf{386.92} & \underline{\textbf{3013.26}} & \stdf{380.32}\\
            BB nodes before opt. & 3.58 & \stdf{0.65} & \underline{\textbf{1.61}} & \stdf{0.68}& 388.68 & \stdf{191.23} & \underline{\textbf{214.31}} & \stdf{127.41} & 2347.24 & \stdf{292.04} & \underline{\textbf{1561.14}} & \stdf{273.65}\\
            $\frac{\mathbb{E}[O_{T^\star}]}{n}$ &-&& 0.97 & \stdf{0.0} &-&   & 0.99 & \stdf{0.0} &-&& 0.99 & \stdf{0.0}\\
            \bottomrule
        \end{tabular}
        }
    }  
\end{table}

As we can see, GCBB achieves better results than classic BB when the graph size grows large. Indeed, the inference of the GCN requires polynomial time in the number of vertices, while the branch and bound is exponential, and, for large instances, the disparity between the two solvers widens as the neural network overhead gets amortized. For graphs of $100$ vertices, the overall hybrid procedure requires less time on average for solving an instance, while for $n=20$ and $n=50$ the total time is smaller for the BB approach. It is worth noting that the metrics concerning branch and bound nodes were already significantly better with instances of size greater than or equal to $50$ cities. We expect this trend to emerge and even amplify with larger instances. To better analyze the differences between the two solvers, we also display performance profiles in \autoref{fig:perf_prof_exact}. In each panel on the left, as time grows, we report the fraction of problem instances that are solved by each algorithm. Meanwhile, in the panel on the right, we also consider instances not yet solved. More precisely, we report the proportion of instances for which each algorithm has reached a superior objective value. We note that, in these latter plots, all lines should approach zero if the algorithms are given enough time. 

As we can observe, when the problem instance is small in size (e.g. 20), the overhead of the neural network (which requires a few seconds to run) is not justified since all the problems are solved by BB in just a few milliseconds. Instances with 50 are instead intermediate ones where the overhead is completely compensated by the quality of the extracted information, and a difference in times is not noticeable. With instances of 100, however, acquiring good heuristic information about the graph instance allows for a significant performance improvement. It's also worth noting that for graphs of these dimensions, not all instances are solved within 10 minutes. 
In this situation, as can be seen from the gap between the curves in \autoref{fig:100-profiles} for 100 cities, the hybrid version terminates execution with a better solution in most cases.

Let us conclude by analyzing the average value assumed by the optimality score $O_T$ on the optimal 1-trees. As we can see, the average score is close to 1, indicating that most of the edges predicted by the neural network as belonging to an optimal tour actually are. 


\section{Conclusion}

This experiment prepares for the findings discussed in the main paper. In this setting, especially compared to Concorde, the information extracted by the network proves useful even for smaller instances, since the underlying solver is less powerful.



\bibliographystyle{elsarticle-harv}
\bibliography{references}

\end{document}